%% file: new_main.tex
\documentclass[letterpaper, 10 pt, conference]{ieeeconf}
\IEEEoverridecommandlockouts

\let\labelindent\relax
\input{configs}

\title{\LARGE \bf Flight Structure Optimization of Modular Reconfigurable UAVs}

\author{Yao Su$^{1}$, Ziyuan Jiao$^{1}$,  Zeyu Zhang$^{1}$, Jingwen Zhang$^{1}$, Hang Li$^{1}$, Meng Wang$^{1}$, Hangxin Liu$^{1\dagger}$
\thanks{This work was supported in part by the National Natural Science Foundation of China (No.52305007, 62376031).}
\thanks{$\dagger$ Corresponding author.}
\thanks{$^{1}$ State Key Laboratory of General Artificial Intelligence, Beijing Institute for General Artificial Intelligence (BIGAI). Emails: \tt{\{suyao, jiaoziyuan, zhangzeyu, zhangjingwen, lihang, wangmeng, liuhx\}@bigai.ai}}
}
\begin{document}
\maketitle

\begin{abstract}
This paper presents a \ac{ga} designed to reconfigure a large group of modular \acp{uav}, each with different weights and inertia parameters, into an over-actuated flight structure with improved dynamic properties. Previous research efforts either utilized expert knowledge to design flight structures for a specific task or relied on enumeration-based algorithms that required extensive computation to find an optimal one. However, both approaches encounter challenges in accommodating the heterogeneity among modules. Our \ac{ga} addresses these challenges by incorporating the complexities of over-actuation and dynamic properties into its formulation. Additionally, we employ a tree representation and a vector representation to describe flight structures, facilitating efficient crossover operations and fitness evaluations within the \ac{ga} framework, respectively. Using cubic modular quadcopters capable of functioning as omni-directional thrust generators, we validate that the proposed approach can (i) adeptly identify suboptimal configurations ensuring over-actuation while ensuring trajectory tracking accuracy and (ii) significantly reduce computational costs compared to traditional enumeration-based methods.
\end{abstract}

\setstretch{0.97}
\section{Introduction}
By docking and undocking with each other, modular \acp{uav} can transform into various structures, exhibiting significant advantages in terms of versatility, robustness, and cost compared to aerial robots with fixed flight configurations~\cite{seo2019modular}. 
An ideal flight structure is typically achieved by human designs leveraging expert knowledge that accommodates task-specific requirements such as payload transportation~\cite{oung2014distributed,mu2019universal}, object manipulation~\cite{saldana2018modquad,saldana2019design,gabrich2018flying,zhao2016transformable}, and dynamic exploration~\cite{xu2022modular,xu2021h}. To mitigate the manual efforts involved in the design process, several algorithms have been proposed to systematically enumerate all possible flight structures and select the optimal one based on a given metric~\cite{chen1998enumerating,liu2010enumeration,gabrich2021finding,feng2022enumerating,stoy2013efficient}. However, as the number of modular \acp{uav} in the system increases, designing an optimal flight structure solely based on human expertise could become infeasible, and the computational complexity of finding such one grows exponentially.

Search sub-configurations with a heuristic to reduce the search space~\cite{gabrich2021finding} and optimization-based methods~\cite{gandhi2020self,xu2023finding} have recently emerged as promising means to improve the computational efficiency in finding optimal or at least near-optimal flight structures of a modular \ac{uav} system. However, when the modules are equipped with different types of sensors, interacting tools, or payloads, the complexity of finding an ideal flight structure through algorithmic solutions escalates because of the \textbf{non-identical weights and inertia parameters} among the modules. Such heterogeneity would significantly influence the dynamical properties of the resulting vehicle, leading to stability and trajectory-tracking challenges. Furthermore, the solution space of finding optimal flight structures with heterogeneous modules becomes too complex for existing methods~\cite{gandhi2020self,gabrich2021finding,xu2023finding}.

\begin{figure}[t!]
\centering
\includegraphics[width=\linewidth]{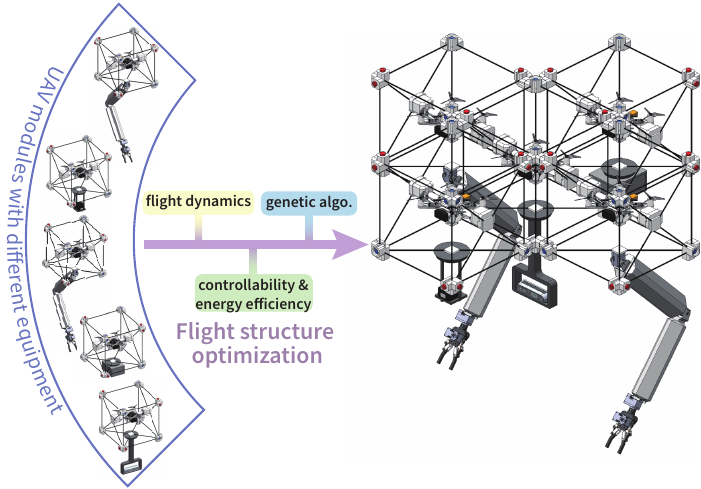}
\caption{\textbf{The optimal structure configuration with five modular \acp{uav} with different installed equipment.} Each module is equipped with either a manipulator, an RGBD camera, a Lidar, or a computing unit, resulting in different weights and inertia parameters. The proposed algorithm efficiently produces an over-actuated flight structure with optimal dynamical properties.}
\label{fig:motiv}
\end{figure}

To tackle the above challenge, we employ \acf{ga} to efficiently find the suboptimal flight structure composed by \ac{uav} modules with different weights and inertia to achieve (i) over-actuation with high trust efficiency and (ii) good controllability for trajectory tracking. The \ac{ga} is a population-based search algorithm capable of addressing both constrained and unconstrained optimization problems through natural selection~\cite{katoch2021review}, thus offering a more flexible formulation to solve the structure optimization problem. To support efficient computation with a minimal budget, we leverage two representations to describe the flight structure of a modular \ac{uav} system within the \ac{ga}: the \textbf{\ac{aim}} (tree representation) for generating new structures with specialized crossover operations~\cite{chen1995determining} and the position vectors for evaluate the fitness of each structure. 

To evaluate the proposed approach, we adopt a customized modular \ac{uav} system. Each module is a quadcopter connected to a cubic docking frame through a 2-\ac{dof} passive gimbal mechanism and thus can be treated as an \textbf{omni-directional thrust generator} after docking with each other horizontally and forming a flight structure~\cite{su2021nullspace,yu2021over}. Specialized equipment with different weights or shapes can also be installed on the bottom of a module's docking frame, resulting in heterogeneous modules. \cref{fig:motiv} illustrates the modular \ac{uav} system and the flight structure found by the proposed algorithm which is over-actuated, \ie independent position and attitude control, and dynamically optimal. 

Our verification first illustrates the convergence process of solving the optimization of flight structures in two cases, each consisting of more than 30 heterogeneous modules. Then, four flight structures at different stages of the optimization process are selected to compare their dynamic properties in trajectory-tracking simulation after implementing a hierarchical controller. Finally, the computational efficiency of the proposed \ac{ga} is studied and contrasted with that of the enumeration-based method. The results highlight that the proposed approach provides dynamically optimal flight structures for large modular \ac{uav} systems with significantly better computational efficiency.

\section{Related Work}\label{sec:related}
Developing \textbf{aerial modular robots} that can adapt to various mission settings flexibly through reconfiguration has attracted a lot of research attention, and progress has been made in modular \ac{uav} system hardware design~\cite{oung2014distributed,mu2019universal,li2021flexibly,saldana2019design,gabrich2020modquad}, dynamics modelling~\cite{zhao2018transformable,yu2021over,nguyen2018novel}, formation control~\cite{saldana2019design}, control allocation~\cite{su2021fast,su2021nullspace,su2024fault}, and docking trajectory planning~\cite{li2019modquad,litman2021vision}. In the above cases, both the structure and module configurations of the system are designed and fixed. More recent work sought to find the optimal/suboptimal flight structure for modular \acp{uav}. Specifically, Gandhi \etal formulated a mixed integer linear programming to reconfigure the flight structure under propeller faults on some modules~\cite{gandhi2020self}. Gabrich \etal proposed a heuristic-based subgroup search algorithm that achieved better computation efficiency than enumerating different types of modules at each module location~\cite{gabrich2021finding}. Xu \etal evaluated a structure's actuation capability by solving an optimization problem~\cite{xu2023finding}. However, the dynamic discrepancy between different modules in terms of weight and inertia parameters is still not considered by these works, prohibiting the fly structure from achieving superior dynamic properties in different task environments.

The problem of \textbf{structure optimization} for a modular robotic system was first investigated by Chen \etal in~\cite{chen1998enumerating}, where an algorithm was proposed to enumerate all non-isomorphic configurations of a modular manipulator system. The following work extended the scope to various modular robot systems with improved computation efficiency~\cite{chen1998enumerating,liu2010enumeration,stoy2013efficient,feng2022enumerating}. However, as the number of modules increases, the set of configurations enlarges factorially, and finding an effective configuration through an exhaustive search becomes infeasible~\cite{chen1995determining,leger1998automated}. On the other hand, utilizing \textbf{\ac{ga}} to find a suboptimal configuration of the modular robotic system has been demonstrated in a simple serial connected structure with much better computation efficiency~\cite{chen1995determining}. In this paper, we generalize this approach to a modular \ac{uav} system with a more complicated \textbf{tree structure}. Taking advantage of our proposed fitness evaluation function and crossover operation for the \ac{ga}, a suboptimal flight structure configuration can be efficiently found. 
     

\section{System Configuration}\label{sec:module}
\begin{figure}[t!]
    \centering
    \includegraphics[width=\linewidth]{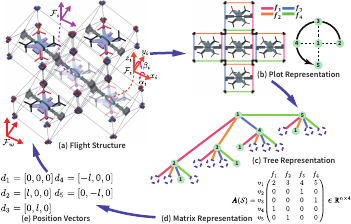}
    \caption{\textbf{Coordinate systems and configuration representations of a flight structure.} Each quadcopter module can connect to four others with its docking faces, and each module serves as an omni-directional thrust generator after connections, making the flight structure over-actuated. Different types of representation of the flight structure are incorporated to support subsequent optimization.}
    \label{fig:structure}   
\end{figure}

Each modular \ac{uav} is constructed by connecting a customized quadcopter to the cubic docking frame (size $l$) by a 2-\ac{dof} passive gimbal mechanism that has no limits in rotation angle. As shown in \cref{fig:structure}(a), a flight structure can be formed with several modules (indexed by $i=1,\cdots,n$) by docking with each other. As each quadcopter module can be utilized as an omni-directional thrust generator, the resulting multirotor flight structure has the potential of being over-actuated to gain better maneuverability with independent position and orientation tracking capabilty~\cite{yu2023compensating}. To represent the entire flight structure, we first index the four docking faces of one module by $\left\{f_1,f_2,f_3,f_4\right\}$ (see \cref{fig:structure}(b)). Then we derive a simplified representation of the flight structure using a matrix (tree) for the subsequent computation applied to the rest of this paper.

\subsection{System Frames Definition \& Notation}
Let $\mathcal{F}_W$ denote the world coordinate frame and $\mathcal{F}_\mathcal{S}$ be the structure frame attached to the geometric center of the flight structure. We define the position of the structure as $\pmb{X}_\mathcal{S}=[x_\mathcal{S},\,y_\mathcal{S},\,z_\mathcal{S}]^\mathsf{T}$, the attitude of the structure in the roll-pitch-yaw convention as $\pmb{\Theta}_\mathcal{S}=[\phi_\mathcal{S},\,\theta_\mathcal{S},\,\psi_\mathcal{S}]^\mathsf{T}$, and the angular velocity $\pmb{\Omega}_\mathcal{S}=[p_\mathcal{S},\,q_\mathcal{S},\,r_\mathcal{S}]^\mathsf{T}$. The frames $\mathcal{F}_{i}$s are attached to the center of the $i$-th module. $\pmb{d}_i=[x_i,\,y_i,\,0]^\mathsf{T}$ denotes the vector from $\mathcal{F}_{i}$ to  $\mathcal{F}_\mathcal{S}$.

\subsection{Flight Structure Configuration}
\label{sec:config}
The configuration of the flight structure refers to a set of
points ($\pmb{d}_i$) in $\mathcal{S}$ that represents the positions of $n$ docked modules w.r.t the structure frame $\mathcal{F}_\mathcal{S}$. Following the idea introduced by Chen \etal~\cite{chen1995determining}, we describe the flight structure as a tree where the module $\textbf{1}$ is chosen as the root. The structure configuration can then be characterized by the \acf{aim}.

Using the configuration in \cref{fig:structure}(a) as an example, the corresponding \ac{aim} is shown in \cref{fig:structure}(d), where each row represents a module, while each column represents a docking face. 
\begin{equation}
    A_{ip}=j, A_{jq}=i, p, q\in\{1, 2, 3, 4\} ,
    \label{eq:condition}
\end{equation} indicates the $p$-th face of module $i$ is connected to the $q$-th face of module $j$. Of note, $\pmb{A}(\mathcal{S})$ is a normalized representation that is independent of module dimension $l$. 

We can calculate $\pmb{d}_i$ from $\pmb{A}(\mathcal{S})$ with the following steps (refers as \textbf{PosTreeSearch} function in the rest of the paper) : 
\begin{enumerate}[leftmargin=*,noitemsep,nolistsep,topsep=0pt]
    \item assuming the module $\textbf{1}$ is at the origin $\pmb{d}_1=[0,0,0]^\mathsf{T}$ and determining the position of each module $\pmb{d}_i$ recursively with the Depth-First tree traversal algorithm and the \textbf{Step} matrix that describes the relative position between two modules according to the docked face:
    \begin{equation}
    \textbf{Step}=\begin{pNiceArray}{cccc}[first-row,first-col]
     & f_1 & f_2 & f_3 & f_4 \\
     x&  1 & 0 & -1 & 0 \\
     y& 0 & 1 & 0 & -1\\
     z& 0 & 0 & 0 & 0
    \end{pNiceArray}\times l\in\mathbb{R}^{3\times4}.
    \label{eq:step}
    \end{equation} 
    Specifically, if $\pmb{d}_i$ is known and \cref{eq:condition} satisfied, then the position of the module $j$ is calculated as:
    \begin{equation}
        \pmb{d}_j=\pmb{d}_i+\textbf{Step}(:,p).
    \end{equation} 
    \item calculating the geometric center position of the structure $\mathcal{S}$ with $\pmb{d}_o=\frac{1}{M}\sum_{i=1}^{n} m_i\pmb{d}_i$, where $M=\sum_{i=1}^{n} m_i$ is the total mass of the flight structure with $m_i$ represents the mass of module $i$;
    \item shifting the geometric center of $\mathcal{S}$ to the origin with $\pmb{d}_i=\pmb{d}_i-\pmb{d}_o$, which ensures $\frac{1}{M}\sum_{i=1}^{n} m_i\pmb{d}_i=0$.
\end{enumerate}

The reason for having both position vectors $\pmb{d}_i$  and \ac{aim} as two representations for the flight structure is twofold: (i) the dynamic property evaluation of each flight structure requires $\pmb{d}_i$, but checking the feasibility of forming a connected structure with $\pmb{d}_i$ of each module can be a tedious process~\cite{chen1998enumerating}; (ii) describing the flight structure as a tree with the AIM representation naturally ensures all the modules are connected. Additionally, dividing the tree into two sub-trees and reconnecting them is an efficient way to generate new structures. This approach simplifies the feasibility check process as only the overlap between modules needs to be considered. Therefore, during flight structure optimization, we maintain two representations of each structure to efficiently generate new structures and evaluate their dynamic properties.  

\begin{figure*}[t!]
        \centering
        \includegraphics[width=\linewidth]{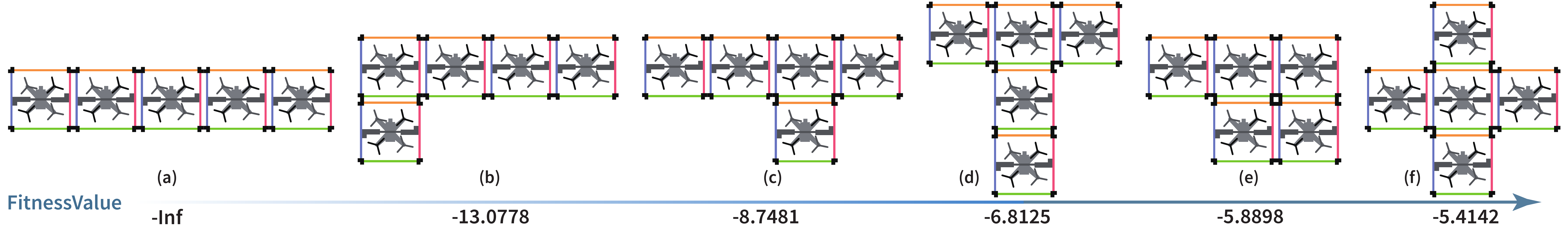}
        \caption{\textbf{The fitness values for different flight structures composed of five same-weighted modules.} Configuration (a) has $-\text{Inf}$ fitness value as it is under-actuated, while (f) has the maximum fitness value (optimal dynamics property) due to its symmetric configuration.}
        \label{fig:fit}
\end{figure*}

\subsection{Flight Structure Dynamics}
Given a flight structure, its translational dynamics can be described as~\cite{ruan2023control}:
\begin{equation}
    M\,\pmb{\ddot{X}}_\mathcal{S} 
    = 
    \prescript{W}{S}{\pmb{R}} (\sum_{i=1}^{n} \prescript{S}{i}{\pmb{R}}\,T_i\pmb{\hat{z}})
   + 
   M\,g\pmb{\hat{z}}, 
    \label{eq:simple_tran}
\end{equation}
where $\pmb{\ddot{X}}_\mathcal{S}$ is the linear acceleration of the flight structure, $g$ is the gravitational acceleration, $\prescript{S}{i}{\pmb{R}}$ is a function of tilting and twisting angles ($\alpha_i$ and $\beta_i$) of $i$th module, and $T_i$ refers to the magnitude of thrust generated by $i$th module, $\hat{\pmb{z}} = \left[0,\,0,\,1\right]^\mathsf{T}$.

Its rotational dynamics can be described as~\cite{su2023sequential}:
\begin{equation}
    \pmb{J}_\mathcal{S}\, \pmb{\dot{\Omega}}_\mathcal{S} = -\pmb{\Omega}_\mathcal{S} \times \pmb{J}_\mathcal{S}\pmb{\Omega}_\mathcal{S}+
 \sum_{i=1}^{n} (\pmb{d}_i \times \prescript{S}{i}{\pmb{R}} \,T_i \pmb{\hat{z}}), 
    \label{eq:simple_rota}
\end{equation}
where $\pmb{J}_\mathcal{S}$ is the total inertia matrix of the structure:
\begin{equation}
    \pmb{J}_\mathcal{S}=\sum_{i=1}^{n}\pmb{J}_i+\sum_{i=1}^{n} m_i
     \begin{bmatrix}
        y_i^2 & -x_iy_i & 0\\
        -x_iy_i &  x_i^2 & 0\\
        0 & 0 & x_i^2+y_i^2
     \end{bmatrix},
     \label{eq:Js}
\end{equation}
with $\pmb{J}_i=Diag(J_{ix},\,J_{iy},\,J_{iz})$ is the inertia matrix of each the module, $\pmb{\dot{\Omega}}_\mathcal{S}$ is the angular acceleration of the structure.

Taking together, the complete dynamics model of the flight structure is:
\begin{equation}
    \begin{bmatrix}
         \pmb{\ddot{X}}_\mathcal{S}\\
         \pmb{\dot{\Omega}}_\mathcal{S}\\
    \end{bmatrix} = 
    \begin{bmatrix}
       {\frac{1}{M}}\prescript{W}{S}{\pmb{R}} &0\\
       0 & \pmb{J}_\mathcal{S}^{-1}
    \end{bmatrix}\pmb{u}
    + 
    \begin{bmatrix}
        g \pmb{\hat{z}} \\
        -\pmb{J}_\mathcal{S}^{-1}(\pmb{\Omega}_\mathcal{S} \times \pmb{J}_\mathcal{S}\pmb{\Omega}_\mathcal{S}) \\
    \end{bmatrix},
    \label{eq:simple_dyna}
\end{equation}
where $\pmb{u}$ is the 6-\ac{dof} wrench generated by all modules with  
\begin{equation}
\label{eq:bu_u}\resizebox{0.91\linewidth}{!}{
$\begin{aligned}
\pmb{u} &= 
    \begin{bmatrix}
    \pmb{u}_{X}\\
    \pmb{u}_{\Omega}
    \end{bmatrix}
    =
    \begin{bmatrix}
        \sum_{i=1}^{n} \prescript{S}{i}{\pmb{R}}\, T_i \pmb{\hat{z}} \\
        \sum_{i=1}^{n} (\pmb{d}_i \times \prescript{S}{i}{\pmb{R}}\, T_i \pmb{\hat{z}}) \\
    \end{bmatrix}
    =    
    \begin{bmatrix}
        \pmb{J}_{X}(\pmb{\alpha},\pmb{\beta})\\\pmb{J}_{\Omega}(\pmb{\alpha},\pmb{\beta})
    \end{bmatrix}
    \pmb{T}, \\
    \pmb{\alpha}&=[\alpha_1,\cdots,\alpha_n]^\mathsf{T}, \pmb{\beta}=[\beta_1,\cdots,\beta_n]^\mathsf{T}, \pmb{T}=[T_1, \cdots,T_n]^\mathsf{T}.  \end{aligned}$}
\end{equation}
Of note, due to the passive gimbal mechanism design and the low-level control of each module eliminating rotating torques in the z-axis~\cite{yu2021over}, there is \textbf{no rotation-induced torque} of propellers pass to the mainframe. Therefore, in the simplified dynamics model \cref{eq:simple_dyna}, the input of each module is only considered as a 3-axis force vector. Besides, the total inertia matrix is approximated as a constant matrix.
\subsection{Force Decomposition-based Analysis}
\label{sec:allo}
Utilizing the force decomposition method, we can transform the nonlinear relationship in \cref{eq:bu_u} to a linear one by defining $\pmb{F}$ as an intermediate variable~\cite{su2024fault,yu2021over}:
\begin{equation}
    \pmb{F}=     
    \begin{bmatrix}    
        F_1\\
        \vdots\\
        F_n\\ 
    \end{bmatrix}\in\mathbb{R}^{3n\times1},
    F_i
    =
    \begin{bmatrix}    
        \sin\beta_{i}\\
        -\sin\alpha_{i}\cos\beta_{i}\\
        \cos\alpha_{i}\cos\beta_{i}\\ 
    \end{bmatrix}T_i,
\end{equation}
Then, the 6-\ac{dof} wrench $\pmb{u}$ can be rewritten as:
\begin{equation}
    \pmb{u}^d = 
    \begin{bmatrix}
        \pmb{J}_{X}(\pmb{\alpha},\pmb{\beta})\\\pmb{J}_{\Omega}(\pmb{\alpha},\pmb{\beta})
    \end{bmatrix}
    \pmb{T}
    = 
    \pmb{WF},
    \label{eq: static}
\end{equation}
where $\pmb{W}\in\mathbb{R}^{6\times3n}$ is a constant allocation matrix with full row rank. Treating $\pmb{F}$ as inputs to the system, its dynamics can be analyzed from a linear perspective. The real inputs tilting angle $\alpha_i$, twisting angle $\beta_i$, and thrust force $ T_{i}$ of each module are computed from $\pmb{F}$ using inverse kinematics.

\section{Flight Structure Optimization}\label{sec:optimization}
To find the optimal flight structure that is over-actuated and dynamically optimized, we first isolate the configuration-related factors from the dynamics equations (allocation matrix $\pmb{W}$ and total inertia matrix $\pmb{J}_s$). Then we design an evaluation function of each flight structure from the dynamics/control perspective. Finally, a \ac{ga} is implemented to find the suboptimal flight structure with improved computation efficiency. 

\subsection{Dynamic Property Evaluation}
The allocation matrix $\pmb{W}$ is derived as~\cite{gabrich2021finding}:
\begin{equation}
\begin{split}
    \pmb{W}=\pmb{\Bar{P}}\pmb{\Bar{R}}=
    \begin{bmatrix}
        \pmb{I}_3 &\cdots& \pmb{I}_3\\
        \pmb{\hat{d}}_1 &\cdots& \pmb{\hat{d}}_n
    \end{bmatrix}
    \begin{bmatrix}
    \prescript{S}{1}{\pmb{R}} &\cdots & 0\\
    &\ddots&\\
    0&\cdots&\prescript{S}{n}{\pmb{R}}
    \end{bmatrix},
\end{split}
\end{equation}
where $\pmb{\Bar{P}}\in\mathbb{R}^{6\times3n}$,
$\pmb{\Bar{R}}\in\mathbb{R}^{3n\times3n}$,
$\pmb{\hat{d}}_i$ is the skew symmetric matrix of $\pmb{d}_i$. From the formulation of $\pmb{\Bar{P}}$, we can easily find that the translational dynamics (the first three rows of $\pmb{W}$) are independent of the configuration. Therefore, we only focus on the rotational dynamics (the last three rows of $\pmb{W}$) in this optimization problem. Besides, the total inertia $\pmb{J}_\mathcal{S}$ is also related to the structure configuration (see \cref{eq:Js}) which is not neglectable. Therefore, we define a matrix $\pmb{\Bar{D}}$ as: 
\begin{equation}
    \pmb{\Bar{D}}=
    \pmb{J}_\mathcal{S}^{-1}\pmb{\hat{D}}=
    \pmb{J}_\mathcal{S}^{-1}
    \begin{bmatrix}
    \pmb{\hat{d}}_1 &\cdots& \pmb{\hat{d}}_n
    \end{bmatrix}.
    \label{eq:dhat}
\end{equation} 
Calculating the required thrust energy index $\lVert\pmb{T}\rVert^2$ and neglecting $\pmb{\Omega}_\mathcal{S} \times \pmb{J}_\mathcal{S}\pmb{\Omega}_\mathcal{S}$, we have:  
\begin{equation}
\begin{split}
    \lVert\pmb{T}\rVert^2
    = 
    \lVert\pmb{F}\rVert^2 
    &= 
    \pmb{u}_{\Omega}^\mathsf{T}\pmb{\hat{D}}^{\dagger T}\pmb{\Bar{R}}\pmb{\Bar{R}}^\mathsf{T}\pmb{\hat{D}}^{\dagger}\pmb{u}_{\Omega}
    =
   \pmb{\dot{\Omega}}_\mathcal{S}^\mathsf{T}\pmb{\Bar{D}}^{\dagger T}\pmb{\Bar{D}}^\dagger \pmb{\dot{\Omega}}_\mathcal{S} \\
   &\leq {\sigma}_\text{max}(\pmb{\Bar{D}}^\dagger)^2\lVert\pmb{\dot{\Omega}}_\mathcal{S}\rVert^2 \quad \forall \pmb{\dot{\Omega}}_\mathcal{S},
    \label{eq:obj2}
\end{split}
\end{equation}
where $(\cdot)^\dagger$ is the Moore–Penrose inverse of a matrix, and ${\sigma}_\text{max}(\cdot)$ is the maximum singular value of a matrix.

Based on the above analysis, we formulate the flight structure evaluation function as :  
\begin{equation}
    \operatorname*{argmax}_{\pmb{\Bar{D}}}  \quad
    -\lambda_1 \text{cond}(\pmb{\Bar{D}})-\lambda_2 {\sigma}_\text{max}(\pmb{\Bar{D}}^\dagger)^2,
\label{eq:fitness}
\end{equation}
where $\text{cond}(\cdot)$ is the condition number of a matrix.  The left half of the objective function $ -\text{cond}(\pmb{\Bar{D}})$ considers the controllability where the full-actuation constraint is implicitly included ($\text{cond}(\pmb{\Bar{D}})=\text{Inf}$ if $\text{rank}(\pmb{\Bar{D}})<3$); the right half $-{\sigma}_\text{max}(\pmb{\Bar{D}}^\dagger)^2$ characterizes the thrust minimization criteria (\cref{eq:obj2}), $\lambda_{1-2}$ are the weighting parameters.

\begin{figure*}[ht!]
    \centering
    \begin{subfigure}[b]{\linewidth}
        \centering
        \includegraphics[width=\linewidth]{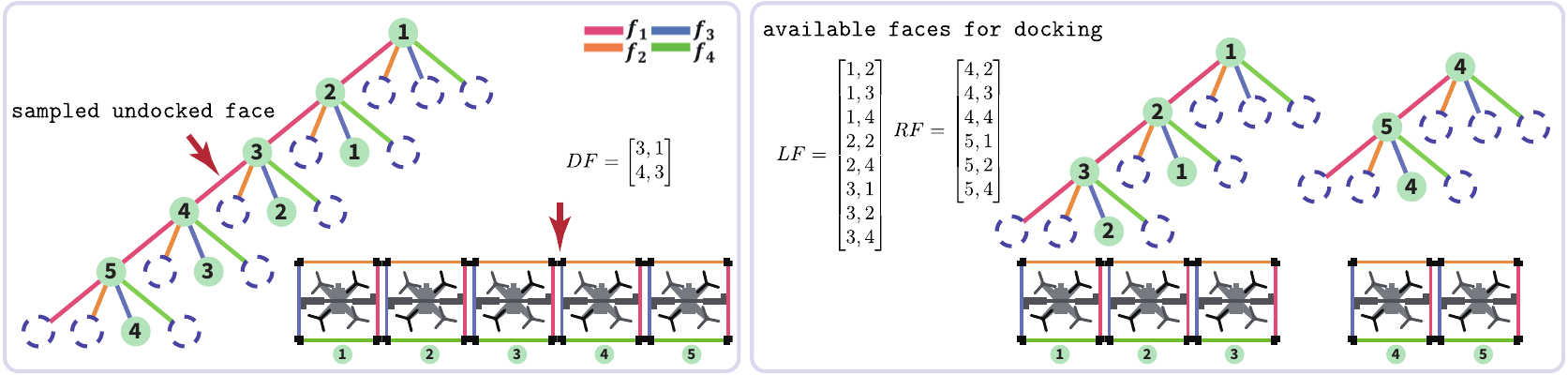}
        \caption{Breaking the initial configuration into two sub-trees.}
        \label{fig:crossover_a}
    \end{subfigure}%
    \\
    \begin{subfigure}[b]{\linewidth}
        \includegraphics[width=\linewidth]{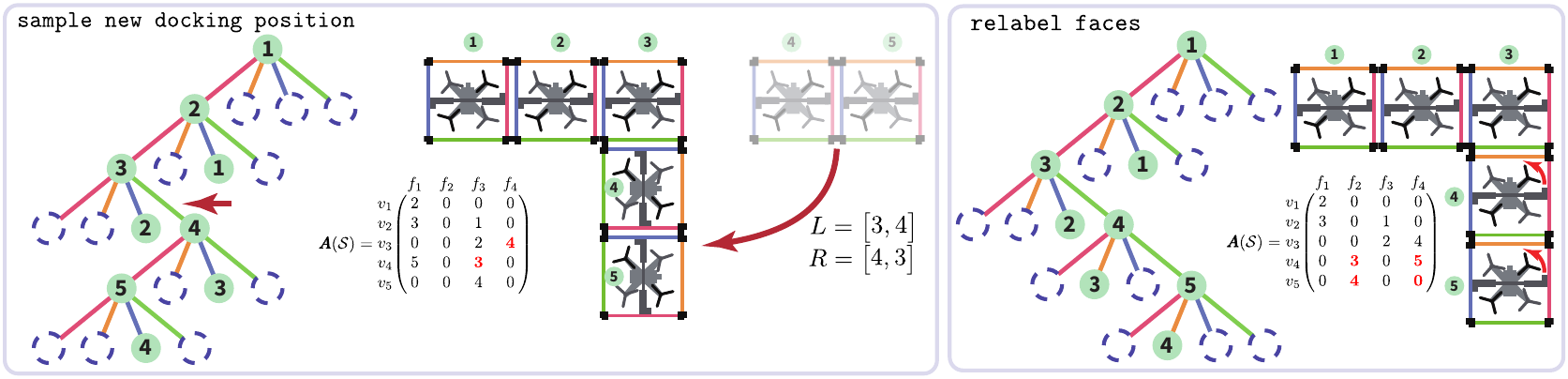}
        \caption{Forming a new configuration by connecting them with a new label assignment.}
        \label{fig:crossover_b}
    \end{subfigure}%
    \caption{\textbf{Steps of a crossover operation.} The crossover operation divides the original tree into two small trees and reconnects them to build a new one. Through the crossover operation, all the feasible configurations can be acquired.} 
        \label{fig:crossover}
\end{figure*}

\subsection{Genetic Algorithm}

Given the number of modules $n$, the total number of different structures can be analytically calculated~\cite{white1969introduction}, and the method of enumerating all feasible $\pmb{A}$ to find an optimal flight structure has been introduced in~\cite{chen1998enumerating}. However, as $n$ increases, the number of $\pmb{A}$ grows factorially and thus it demands excessive computation. To overcome the limitation of existing methods, we propose a \ac{ga} to find a suboptimal flight structure configuration with better efficiency. Different from the serially connected structure handled by~\cite{chen1995determining}, we implement the \ac{ga} (see \cref{alg:GA}) to deal with more complicated tree representation in this work.

At first, each chromosome in the population is randomly initialized as a serial chain, \eg a chain with five modules in \cref{fig:fit}a. In each generation, (i) we first evaluate the fitness of the population; then (ii) we pick some chromosome from the population through a tournament; (iii) every picked chromosome is utilized to generate new children by crossover operation; (iv) we evaluate the existing population along with their children and select new population from them as the next generation. The fitness evaluation and crossover operation as two main components will be introduced in detail next, which correspond to \textbf{computeFitness} (\cref{alg:fitness}) and \textbf{Crossover} (\cref{alg:crossover}), respectively.

\begin{algorithm}[t!]
    \fontsize{8pt}{8pt}\selectfont
    \caption{\textbf{Genetic Algorithm}} 
    \label{alg:GA}  
    \SetKwInOut{KIN}{Input}
    \SetKwInOut{KOUT}{Ouput}
    \SetKwInOut{PARAM}{Params}
    \KIN{Number of \ac{uav} modules: $n$,\\
    Mass of each module: $m_{1..n}$,\\
    Inertia of each module: $J_{1..n}$
    }
    \KOUT {Optimized structure: $A^*$
    }
    \PARAM{Maximum population size: $\text{PopSize}$,\\
    Maximum generations: $\text{GSize}$, \\
    Tournament size: $\text{TSize}$, \\
    Number of children: $\text{CSize}$, \\
    Crossover probability: $\text{CrossP}$,\\
    Number of generations for convergence check: $K$
    }
    \textcolor{blue}{\tcp*[h]{Initialize population and fitness}}\\
    $\text{Pop}\gets{\text{Initialize}(\text{PopSize},n)}\in\mathbb{R}^{n\times4\times{\text{PopSize}}}$\;
    $\text{Fit} \gets \text{computeFitness}(\text{Pop})$; see \cref{alg:fitness}\\
    \textcolor{blue}{\tcp*[h]{Generation iteration}}\\
    \For{$G_i\in1\cdots \text{GSize}$}{
        $\text{NewPop}\gets{\text{Pop}}$\; 
        \For{$i\in1 \cdots \text{TSize}$}{
            \textcolor{blue}{\tcp*[h]{Tournament selection}}\\
            $\text{idx}\gets \text{TSelect}(\text{Pop},\text{Fit},\text{TSize})$\; 
            $\text{Chromo} \gets \text{Pop}(:,:,\text{idx})$\; 
            \textcolor{blue}{\tcp*[h]{Generate children}}\\
            \For{$j\in1 \cdots \text{CSize}$}{ 
                \eIf{$rand < \text{CrossP}$}{    
                 $\text{NewPop}(:,:,\text{end}+1) \gets \text{Crossover}(\text{Chromo})$\\}
                {$\text{NewPop}(:,:,\text{end}+1) \gets \text{Chromo}$\;}
            }
        }
               
            $\text{NewFit}\gets \text{computeFitness}(\text{NewPop},m_{1..n},J_{1..n})$\;
            \textcolor{blue}{\tcp*[h]{Select new population}}\\
            $\text{Pop}, \text{Fit}\gets \text{PopSelect}(\text{NewPop},\text{NewFit},\text{PopSize})$\;
            \If{$\text{max}(\text{Fit})$ repeat for $K$ generations}{break \textcolor{blue}{\tcp*[h]{Optimization converges}}}
    }
    $A^* \gets \text{Pop}(:,:,1)$\; 
\end{algorithm}

\begin{algorithm}[t!]
    \fontsize{8pt}{8pt}\selectfont
    \caption{\textbf{computeFitness}}
    \label{alg:fitness}  
    \SetKwInOut{KIN}{Input}
    \SetKwInOut{KOUT}{Ouput}
    \SetKwInOut{PARAM}{Params}
    \KIN{$\text{Pop}$, $m_{1..n}$, $J_{1..n}$}
    \KOUT{$\text{Fitness}$}
    \PARAM{Quadcopter size in \cref{eq:step}: $l$ \\
    The root of tree representation: $\text{root}$\\
    Weighing parameters: $\lambda_1,\lambda_1$}
    $\text{Fitness}\gets\emptyset$\;
    \For{$\text{Chromo}\in {\text{Pop}}$}{
    $d \gets \text{PosTreeSearch}(\text{root},\text{Chromo},\text{Step})$\;
    $Js \gets $\cref{eq:Js}, $D \gets $\cref{eq:dhat}\;
    $\text{FValue}\gets-\lambda_1\text{cond}(D)-\lambda_2\text{max}(\text{svd}(D))^2$\;
    $\text{Fitness}\gets \text{Fitness}\cup \{\text{FValue}\}$}
\end{algorithm}

\begin{algorithm}[t!]
    \fontsize{8pt}{8pt}\selectfont
    \caption{\textbf{Crossover}} 
    \label{alg:crossover}  
    \SetKwInOut{KIN}{Input}
    \SetKwInOut{KOUT}{Ouput}
    \SetKwInOut{PARAM}{Params}
    \KIN{$\text{Chromo}$}
    \KOUT{$\text{NewChromo}$}
    $\text{Feasible} \gets \text{False}$\;     
    \While{$\neg \text{Feasible}$}{
      $ChR \gets \text{Chromo}, r \gets \text{randomInt}(1,n)$\;
      
       \textcolor{blue}{\tcp*[h]{Find feasible docking faces, and current docking face pairs}}\\
      $[LF,RF,DF]\gets \text{DFSTreeSearch}(r,ChR)$\;
      \textcolor{blue}{\tcp*[h]{Delete current docking connections}}\\
      $ChR(DF(1,:))\gets0$, $ChR(DF(2,:))\gets0$\;
      \textcolor{blue}{\tcp*[h]{Random select a connecting face L}}\\
      $L\gets LF(\text{randomInt}(1,\text{length}(LF)),:)$\;
      \textcolor{blue}{\tcp*[h]{Random select a connecting face R}}\\
      $R \gets RF(\text{randomInt}(1,\text{length}(RF)),:)$\;
      \textcolor{blue}{\tcp*[h]{Connect as a new tree}}\\
      $ChR(L) \gets R(1,1)$, $ChR(R) \gets L(1,1)$\;
      \textcolor{blue}{\tcp*[h]{Rotate small tree}}\\
      $ChR \gets \text{RotateFaces}(ChR,RF)$\;
      \textcolor{blue}{\tcp*[h]{Check new tree with no overlap}}\\
      $\text{Feasible} \gets \text{CheckFeasible}(ChR)$\;
      }
      $\text{NewChromo} \gets ChR$\;
\end{algorithm}

\textbf{Fitness evaluation:} 
To evaluate the optimality of different structure configurations, we choose \cref{eq:fitness} as the fitness function, which requires both the position vector $\pmb{d_i}$ and total inertia matrix $\pmb{J}_\mathcal{S}$ derived from the \ac{aim} of a structure for evaluation. Following the method introduced in \cref{sec:config}, the position of each module is first decided with the Depth-First tree traversal algorithm. And then $\pmb{\Bar{D}}$ matrix is built with \cref{eq:dhat,eq:Js} for fitness value calculation. The detail of this function is described in \cref{alg:fitness}. In \cref{fig:fit}, some structure configurations with five same-weighted modules are plotted and evaluated with our proposed fitness function as examples. We can easily find the \cref{fig:fit}(a) has the minimum fitness value due to under-actuation, while plus shape configuration \cref{fig:fit}(f) has the maximum fitness value. The fitness function will be studied and discussed in \cref{sec:sim}.

\textbf{Crossover operation:}  As demonstrated in \cref{fig:crossover}, we define the crossover operation as breaking the tree structure into two subtrees and reconnecting them into a new tree. For a given configuration input, (i) we first randomly pick a module $r$ as the breaking point, and utilize a Depth-First Tree Search algorithm to break the tree into two small trees, then we collect all the possible docking faces on the left tree and the right tree, as $LF$ and $RF$ respectively (see \cref{fig:crossover}(a)); (ii) With one randomly picked element from both sets ($L$ and $R$) as the new pair of docking face, we connect two subtrees into a new one, then we rotate the faces of the small tree to make sure the position relationship is identical to the \text{Step} matrix \cref{fig:crossover}(b) and check the feasibility of the new structure. This function is summarized in \cref{alg:crossover}.

\section{Evaluation}\label{sec:sim}
Through a series of studies, we demonstrated that the proposed method (i) effectively solved the flight structure optimization problem with a large number of modules with different weights, (ii) gradually improved the dynamic properties of an over-actuated flight structure through the proposed crossover operation and fitness evaluation, and (iii) significantly outperformed a traditional emulation-based algorithm in terms of computing time.

\subsection{Simulation Setup}
Utilizing the Simscape module of Matlab Simulink, we developed a simulator where the characteristics of real systems were included, such as control frequencies, motor dynamics, thrust saturation, and measurement noise. Implementing the hierarchical controller developed in our previous work~\cite{yu2021over,su2021nullspace,su2022downwash,su2024real}, we could compare the dynamic properties of generated flight structures by testing their trajectory-tracking performance in simulation.

\subsection{Evaluation Results}

\begin{figure}[t!]
        \centering
        \includegraphics[width=\linewidth,trim=2cm 0cm 0.5cm 0cm,clip]{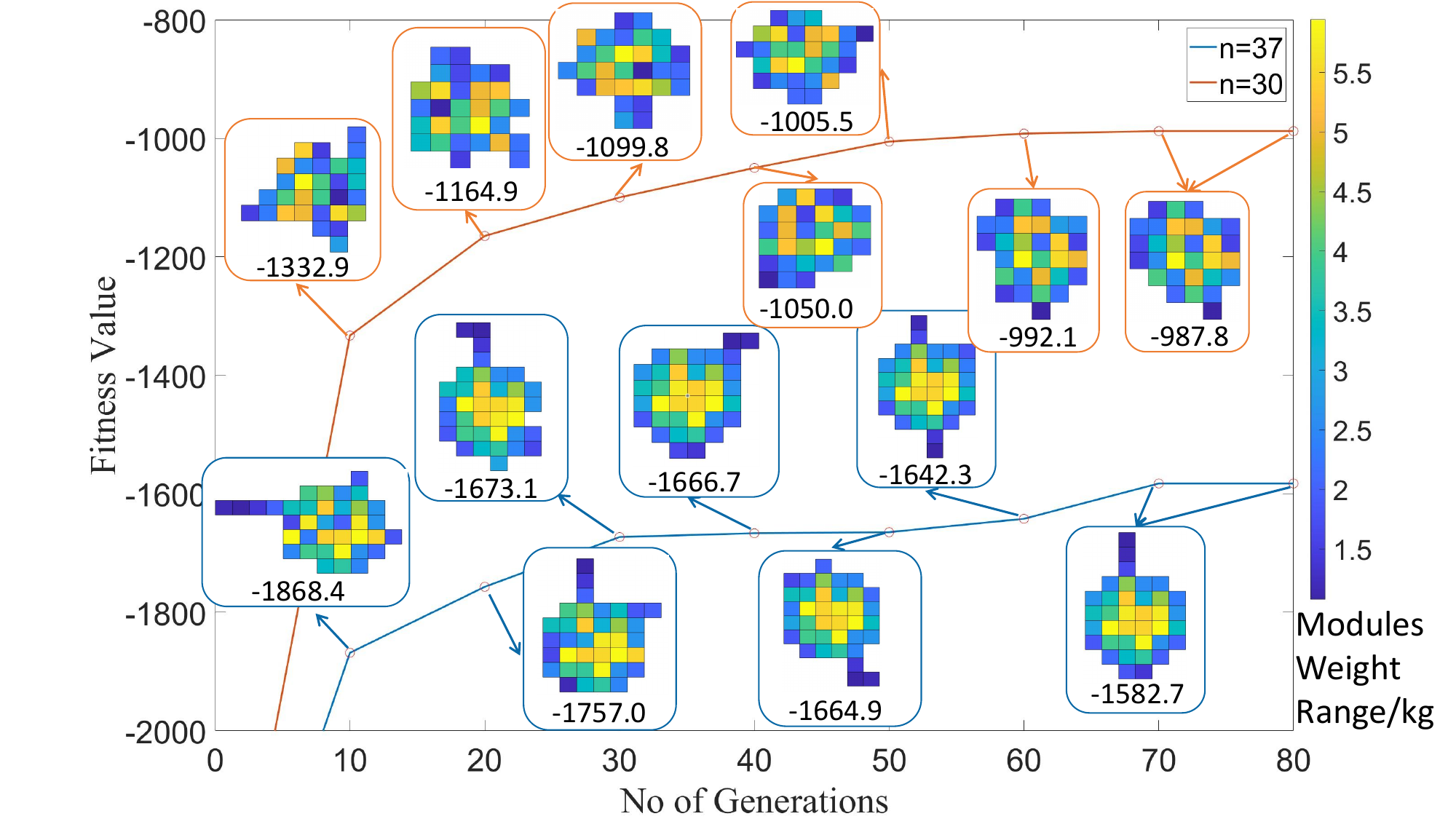}
        \caption{\textbf{The optimization process of generating an optimal flight structure for a large modular UAV system swarm with different-weighted modules.} Two cases with $n=30$ and $n=37$ are presented to illustrate the evolution of the optimization process.} \label{fig:modules37}
\end{figure}

\textbf{Optimizing a complex structure:} In this study, we demonstrated that the proposed \ac{ga} could scale up to optimize flight structure configuration for a large swarm with 30 heterogeneous modules (\textit{Case 1}) and another with 37 modules (\textit{Case 2}). The weight of each module was randomly sampled from $1-5.5~kg$, indicated by the color bar. As seen in \cref{fig:modules37}, During the optimization process, the fitness value for \textit{Case 1} ($n=30$) gradually improved from $-Inf$ and eventually converged to $-1582.7$ after $80$ generations. The final flight structure was over-actuated and formed an approximately symmetric configuration with heavier modules in the middle, which had better controllability across all directions. A similar result could be observed for \textit{Case 2} ($n=37$) as well.

\begin{table}[b!]
    \centering
    \caption{\textbf{Correspondence between structures' fitness value rank and their flying performance}. Four structures (the 10th, 20th, 40th, and 80th generations) that correspond to the \textit{Case 1} of \cref{fig:modules37} are selected to compare their dynamics properties in simulation, and the converged structure indeed performs better than the others.}
    \resizebox{0.83\linewidth}{!}{
    \begin{tabular}{ccccx}
            \toprule
            \textbf{Structure} & \textbf{$G_i(10)$}  & \textbf{$G_i(20)$}  & \textbf{$G_i(40)$}  & \textbf{$G_i(80)$} \\
            \midrule
            Fitness value & $-1332.9$ & $-1164.9$ & $-1050.0$& $-987.8$\\
            \midrule
            $\sum\lVert\pmb{T}\rVert^2(10^7)$& $5.8587$& $5.8521$&$5.8432$& $5.8401$ \\
            Pos RMS error (m)& $ 0.1889$ &$0.1883$ & $0.1878$& $0.1874$ \\
            Att RMS error (rad) & $ 0.1693$& $0.1407$ & $0.1221$& $0.1171$ \\
            \bottomrule
    \end{tabular}}
    \label{tab:config_compare}
\end{table}

\begin{figure}[t!]
        \centering
        \includegraphics[width=0.85\linewidth,trim=0cm 0cm 0cm 0cm,clip]{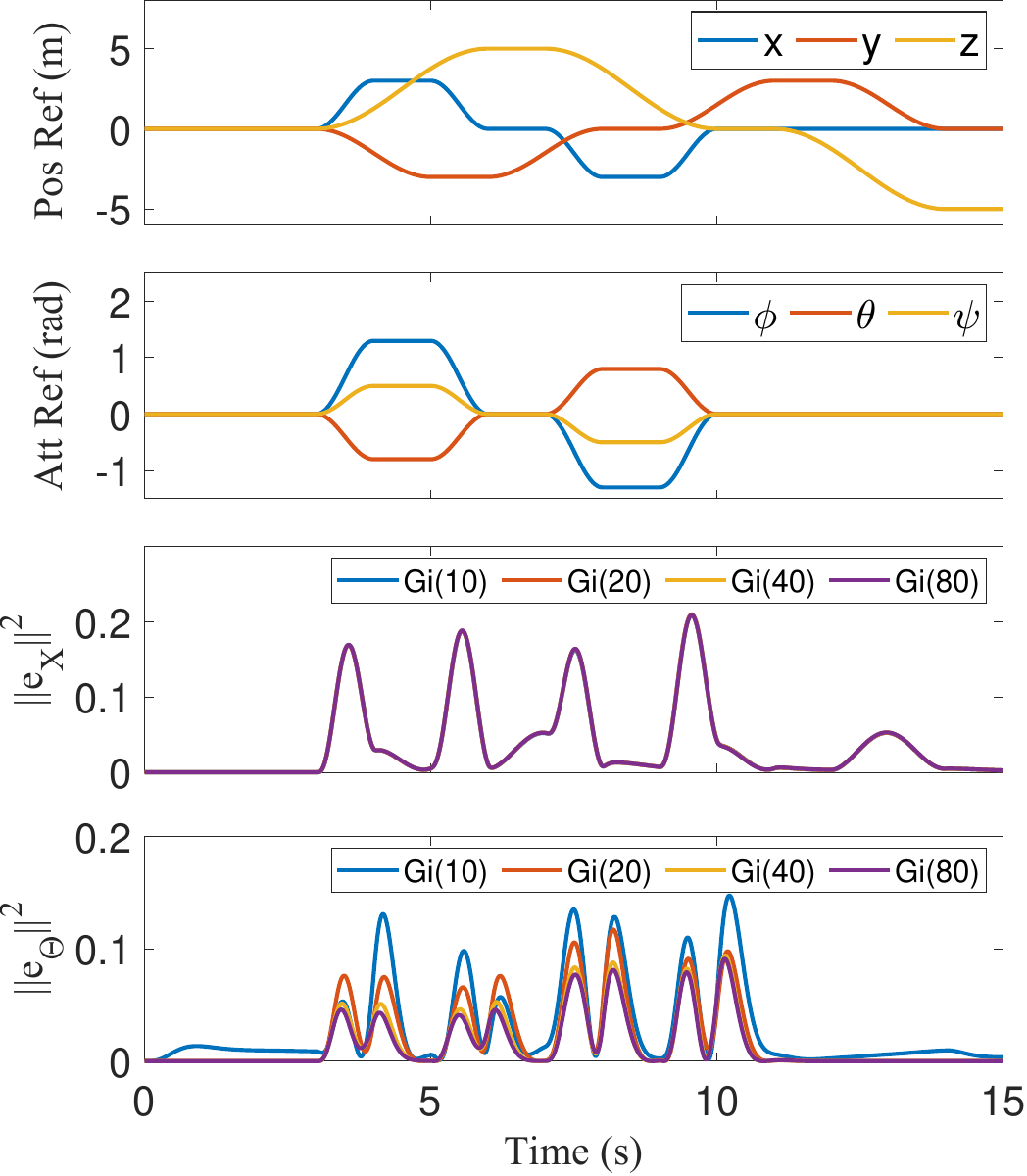}
        \caption{\textbf{Comparison of the tracking performance of different flight structures composed of 30 different-weighted modules.} Although all 4 structures are over-actuated, some outperform others in terms of trajectory tracking accuracy due to better controllability.} 
        \label{fig:compare_five}
\end{figure}

\textbf{Generating efficient flight structure:} Four flight structures (the 10th, 20th,
40th, and 80th generations) generated along the optimization process of \textit{Case 1} in \cref{fig:modules37}, who have an increasing fitness score, were selected to evaluate their dynamics properties in tracking a challenging 6-\ac{dof} trajectory. The thrust energy consumed by the structures to track the trajectory and the corresponding tracking errors were plotted~\cref{fig:compare_five}. \cref{tab:config_compare} further listed the accumulated energy cost derived from thrust and the tracking RMS errors of each structure along the entire trajectory. Despite these 4 flight structures being over-actuated and could independently track position and orientation commands, the converged structure configuration ($G_i(80)$) performed the best in terms of the least energy cost and RMS errors. This configuration also yields the highest fitness value, demonstrating its efficacy in ensuring over-actuation with better dynamics properties.

\begin{figure}[ht!]
\centering
\includegraphics[width=0.9\linewidth,trim=0.5cm 9.4cm 0.5cm 9.4cm,clip]{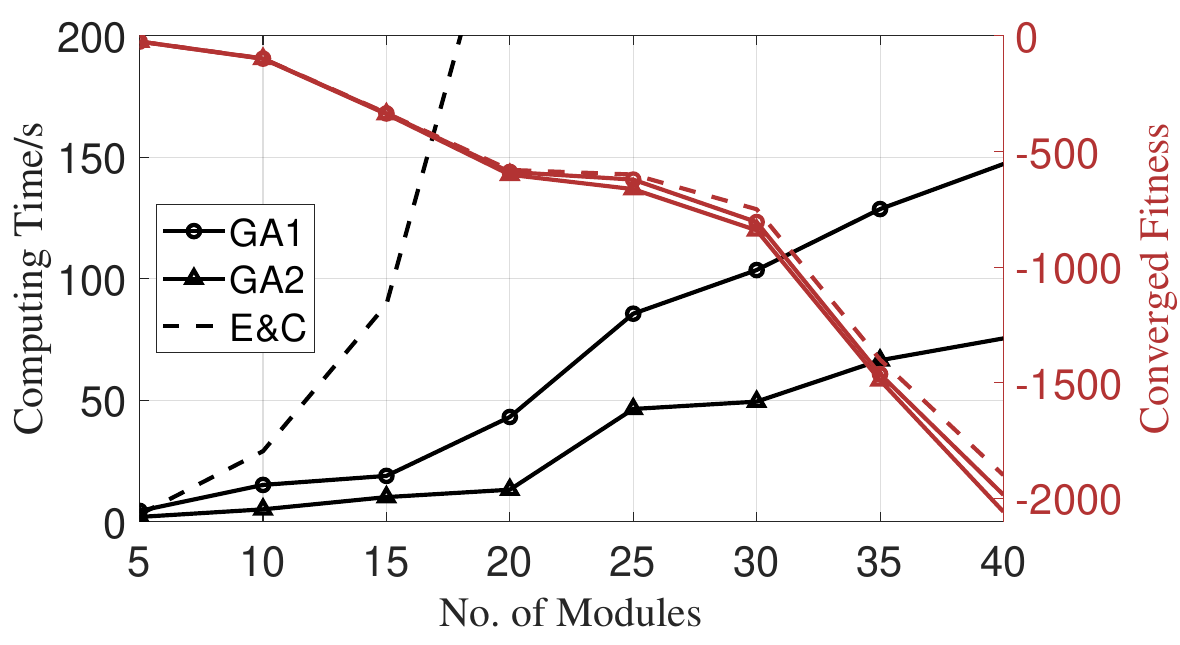}
\caption{\textbf{Computation speed and optimality of the proposed \ac{ga} and the baseline enumeration method.} The \ac{ga} with two population sizes performs significantly faster than that of the baseline. Though the global optimality is not guaranteed, the \ac{ga} could still converge to a suboptimal solution, showing a good trade-off between computation efficiency and optimality.}
\label{fig:speed}
\end{figure}

\textbf{Computation speed:} To demonstrate the proposed method's improvement in computation efficiency, we implemented a classic emulation-based method~\cite{chen1998enumerating,liu2010enumeration} as the baseline and compared its computing time and converged fitness with those of the proposed \ac{ga} with two population sizes ($GA1$ and $GA2$; see \cref{tab:para} for detailed parameters). Using a desktop with AMD Ryzen9 5950X CPU and 64.00 GB RAM, the results are shown in \cref{fig:speed}. The proposed \ac{ga} significantly shortened the required computing time compared with the baseline, especially when the number of modules $n$ became large. Meanwhile, the converged suboptimal configuration found by both settings was close to the global optimal one in terms of fitness. Of note, lowering the population size (\ie $GA1$ \vs $GA2$) could reduce computation, but would sacrifice the optimality, which implies the necessity of balancing optimality and computation efficiency.

\begin{table}[ht!]
    \small
    \centering
    \caption{\textbf{Genetic Algorithm Parameters}}
    \resizebox{0.86\linewidth}{!}{
    \begin{tabular}{lccccccc}
            \toprule
            \textbf{Parameter} & $\text{PopSize}$ & $\text{GSize}$&$\text{TSize}$&$\text{CSize}$&$\text{CrossP}$&$K$ & $Set$ \\
            \midrule
            \multirow{2}{*}{\textbf{Value}} 
             & $3000$&$100$&$100$&$30$&$0.95$&$10$ & $GA1$\\
            & $1000$&$100$&$100$&$30$&$0.95$&$10$  & $GA2$\\
            \bottomrule
    \end{tabular}
    }
    \label{tab:para}
\end{table}

\section{Conclusion}\label{sec:conclusion}
In this paper, we devised a novel optimization method aimed at finding efficient flight structures for modular UAV systems. This method utilized genetic algorithms (GA) to address an energy-minimizing objective function while satisfying over-actuation constraints. By representing the configuration as a tree structure for customized crossover operations and as a position vector for fitness evaluation, the \ac{ga} effectively tackled the optimization problem with significantly reduced computational costs compared to existing emulation-based algorithms for flight structure optimization. Based on our customized modular UAV system, simulation studies illustrated (i) the algorithm's ability to identify efficient flight structures for a large number of modules, (ii) the superior dynamic performance of the resulted flight structure in tracking challenging trajectories, and (iii) the improvement in computational efficiency. Looking ahead, our future plan includes conducting experiments with physical platforms and further enhancing computational capabilities to enable online reconfiguration based on varying task constraints.

{
\bibliographystyle{ieeetr}
\bibliography{reference}
}

\end{document}

%% file: configs.tex
\let\labelindent\relax
\usepackage{graphicx} \graphicspath{ {figures/} }
\usepackage{amsmath,amssymb,mathabx,amsbsy,mathtools,etoolbox}
\usepackage[utf8]{inputenc} 
\usepackage[T1]{fontenc}    

\usepackage{algorithmic}
\usepackage[ruled,vlined]{algorithm2e}

\usepackage{acronym}
\usepackage{enumitem}
\usepackage[breaklinks,colorlinks,linkcolor=black]{hyperref}
\usepackage{balance}
\usepackage{xspace}
\usepackage{setspace}
\usepackage[skip=3pt,font=small]{subcaption}
\usepackage[skip=3pt,font=small]{caption}
\usepackage[dvipsnames,svgnames,x11names]{xcolor}
\usepackage[capitalise,nameinlink]{cleveref}
\usepackage{booktabs,tabularx,colortbl,multirow,multicol,array,makecell}
\usepackage{pifont}
\usepackage{cite}
\usepackage{nicematrix}

\makeatletter
\DeclareRobustCommand\onedot{\futurelet\@let@token\@onedot}
\def\@onedot{\ifx\@let@token.\else.\null\fi\xspace}
\def\eg{\emph{e.g}\onedot} 

\def\ie{\emph{i.e}\onedot}

\def\vs{\emph{vs}\onedot}

\def\etal{\emph{et al}\onedot}
\makeatother

\makeatletter
\def\BState{\State\hskip-\ALG@thistlm}
\makeatother

\makeatletter
\renewcommand{\paragraph}{%
  \@startsection{paragraph}{4}%
  {\z@}{0ex \@plus 0ex \@minus 0ex}{-1em}%
  {\hskip\parindent\normalfont\normalsize\bfseries}%
}
\makeatother

\crefname{algorithm}{Alg.}{Algs.}
\Crefname{algocf}{Algorithm}{Algorithms}
\crefname{section}{Sec.}{Secs.}
\Crefname{section}{Section}{Sections}
\crefname{table}{Tab.}{Tabs.}
\Crefname{table}{Table}{Tables}
\crefname{figure}{Fig.}{Fig.}
\Crefname{figure}{Figure}{Figure}
\crefname{equation}{Eq.}{Eqs.}
\Crefname{equation}{Figure}{Figure}

\definecolor{gblue}{HTML}{D9E6FF}
\definecolor{gred}{HTML}{DB4437}
\definecolor{ggreen}{HTML}{0F9D58}

\definecolor{mygray}{gray}{.92}

\newcolumntype{x}{>{\columncolor{gblue}}c}

\acrodef{qp}[QP]{Quadratic Programming}
\acrodef{aim}[AIM]{assembly incidence matrix}
\acrodef{fd}[FD]{Force Decomposition}
\acrodef{dof}[DoF]{Degree of Freedom}
\acrodef{ros}[ROS]{Robot Operating System}
\acrodef{uav}[UAV]{Unmanned Aerial Vehicle}
\acrodef{dof}[DoF]{Degree-of-freedom}
\acrodef{com}[CoM]{center of mass}
\acrodef{mrr}[MRR]{modular reconfigurable robot}
\acrodef{ga}[GA]{Genetic Algorithm}